\documentclass{egpubl_pg_short}
\usepackage{pg2019s}
 
\WsPaper           

\usepackage[T1]{fontenc}
\usepackage{dfadobe}  

\usepackage{cite}  
\BibtexOrBiblatex
\electronicVersion
\PrintedOrElectronic
\ifpdf \usepackage[pdftex]{graphicx} \pdfcompresslevel=9
\else \usepackage[dvips]{graphicx} \fi

\usepackage{egweblnk} 
\usepackage{times}
\usepackage{epsfig}
\usepackage{graphicx}
\usepackage{amsmath}
\usepackage{amssymb}

\usepackage{enumitem}
\DeclareMathOperator*{\argmax}{arg\,max}
\DeclareMathOperator*{\argmin}{arg\,min}
\usepackage{algorithm}
\usepackage{algorithmic}
\usepackage{hyperref}

\begin{document}

\newcommand{\jb}[1]{{\textbf{\color{blue}[JB: #1]}}}
\newcommand{\bl}[1]{{\color{black}#1}}

\title{LPaintB: Learning to Paint from Self-Supervision}
\author[B. Jia \& J. Brandt \& R. Mech \& B.Kim \& D.Manocha]
{\parbox{\textwidth}{\centering Biao Jia $^{1}$
         Jonathan Brandt $^{2}$
         Radom\'{i}r M\v{e}ch $^{2}$
         Byungmoon Kim $^{2}$
         Dinesh Manocha $^{1}$
        }
        \\
{\parbox{\textwidth}{\centering $^1$ University of Maryland at College Park, U.S.A\\
         $^2$ Adobe Research, U.S.A 
       }
}
\thanks{
This work was supported in part by ARO grant W911NF-18-1- 0313 and Intel.
}
}
\maketitle
\begin{abstract}
We present a novel reinforcement learning-based natural media painting algorithm. Our goal is to reproduce a reference image using brush strokes and we encode the objective through observations. Our formulation takes into account that the distribution of the reward in the action space is  sparse and  training a reinforcement learning algorithm from scratch can be difficult.
We present an approach that combines self-supervised learning and reinforcement learning to  effectively transfer negative samples into positive ones and change the reward distribution. We demonstrate the benefits of our painting agent to reproduce reference images with brush strokes. 
The training phase takes about one hour and the runtime algorithm takes about 30 seconds on a GTX1080 GPU reproducing a  $1000\times800$ image with 20,000 strokes. More details can be found at \url{http://gamma.umd.edu/LPaintV}.
\end{abstract}

\section{Introduction}

\label{sec:intro}


Digital painting systems are increasingly used by artists and content developers for various applications. One of the main goals has been to
simulate popular or widely-used painting styles.
With the development of non-photorealistic rendering techniques, including stroke-based rendering and painterly rendering ~\cite{hertzmann1998painterly, winkenbach1996rendering}, specially-designed or hand-engineered methods can increasingly simulate the painting process by applying heuristics. In practice, these algorithms can generate compelling results, but it is difficult to extend them to new or unseen  styles.

Over the last decade, there has been considerable interest in using machine learning methods for digital painting. These methods include image synthesis algorithms based on convolutional neural networks, including modeling the brush~\cite{xie2012}, generating brushstroke paintings~\cite{xie2015stroke}, reconstructing paintings in specific styles ~\cite{tang2018animated}, constructing stroke-based drawings~\cite{DBLP:journals/corr/HaE17}, etc. Recent developments in generative adversarial networks \cite{goodfellow2014generative} and variational autoencoders~\cite{kingma2013auto} have led to the development of image generation algorithms that can be applied to painting styles~\cite{zhu2017unpaired, zhou2018learning, DBLP:journals/corr/abs-1803-04469, karras2017progressive, sangkloy2017scribbler}.

\begin{figure}[t]
\centering
\includegraphics[width=0.48\textwidth]{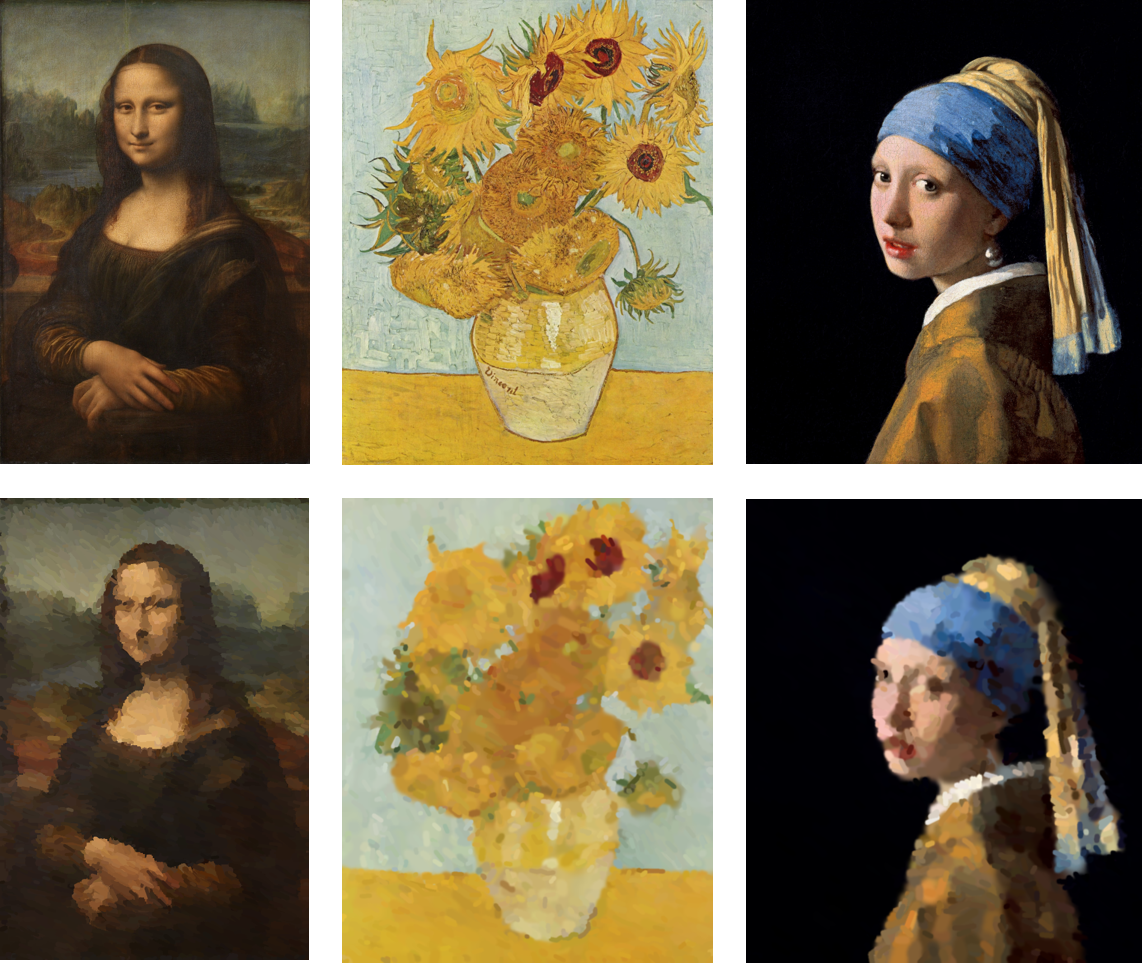}
\caption{\emph{Results Generated by Our Painting Agent:} We use three paintings
(top row) as the reference images to test our novel self-supervised learning algorithm. 
Our trained agent automatically generates the digitally painted image (bottom row) of the corresponding column in about 30 seconds without the need of a paired dataset of human artists. 
}
\label{fig:caption}
\vspace{-25px}
\end{figure}

One of the goals  is to develop an automatic or intelligent painting agent that can develop its painting skills by imitating reference paintings.
In this paper, we focus on building an intelligent painting agent that can reproduce a reference image in an identical or transformed style with a sequence of painting actions. 
Unlike methods that directly synthesize images bypassing the painting process, we focus on a more general and challenging problem of training a painting agent from scratch using reinforcement learning methods. 
 \cite{xie2015stroke, xie2013personal, xie2012, zhou2018learning} also use reinforcement learning to solve the problem.
All the methods encode goal states, which are usually defined as reference images, to the observations. 
This set-up is different from classic reinforcement learning tasks because, while the problem introduces an implicit objective to the policy network of reinforcement learning, the distribution of the reward in the action space can be very sparse and it makes training a reinforcement learning algorithm from scratch very difficult. To solve the problem, \cite{xie2015stroke, xie2013personal, xie2012, zhou2018learning} pre-train the policy network with a paired dataset consisting of images and corresponding actions defined in \cite{xie2012}. 
However, it is very expensive to collect such a paired dataset of human artists and we need to explore other unsupervised learning methods.


\noindent{\textbf{Main Results:}}
We present a reinforcement learning-based algorithm (LPaintB) that incorporates self-supervised learning to train a painting agent on a limited number of reference images without paired datasets. Our approach is data-driven and can be generalized by expanding the image datasets. Specifically, we adopt proximal policy optimization (PPO)~\cite{schulman2017proximal} by encoding the current and goal states as observations and the continuous action space defined based on configurations of the paintbrush like length, orientation and brush size. 
The training component of our method only requires the reference paintings in the desired artistic style and does not require paired datasets collected by human artists.
We use a self-supervised learning method to increase sampling efficiency. 
By replacing the goal state of an unsuccessful episode with its final state, we automatically generate a paired dataset with positive rewards. 
After applying the dataset to retrain the model using reinforcement learning, our approach can efficiently learn the optimal policy.
The novel contributions of our work include:
\begin{itemize}[noitemsep,nolistsep]
\itemsep0em 
    \item An approach for collecting supervised data for painting tasks by self-supervised learning. 
    \item An adapted deep reinforcement learning network that can be trained using human expert data and self-supervised data, though we mostly rely on self-supervised data.
    \item An efficient rendering system that can automatically generate stroke-based paintings of desired resolutions by our trained painting agent.  
\end{itemize}

We evaluate our approach by comparing our painting agent with prior painting agents that are trained  from scratch by reinforcement learning ~\cite{jia2019paintbot}. We collect 1000 images with different color and patterns as the benchmark and compute L2 Loss between generated images and reference images. 
Our results show that self-supervised learning can efficiently collect paired data and can accelerate the training process.
The training phase takes about 1 hour and the runtime algorithm takes about 30 seconds on a GTX 1080 GPU for high-resolution images.



\section{Related Work}
In this section, we give a brief overview of prior work on non-photorealistic rendering and the use of machine learning techniques for image synthesis.
\subsection{Non-Photorealistic Rendering}
Non-photorealistic rendering methods render a reference image as a combination of strokes by determining many properties like position, density, size, and color. 
To mimic the oil-painting process, Hertzmann~\shortcite{hertzmann1998painterly} renders the reference image into primitive strokes using gradient-based features. 
To simulate mosaic decorative tile effects, Hauser~\shortcite{hausner2001simulating} segments the reference image using Centroidal Voronoi diagrams. Many algorithms have been proposed for specific artistic styles, such as stipple drawings~\cite{deussen2000floating}, pen-and-ink sketches~\cite{salisbury1994interactive} and oil paintings~\cite{zeng2009image} ~\cite{lindemeier2015hardware}.
The drawback of non photo-realistic rendering methods is the lack of generalizability to new or unseen styles. Moreover, they may require hand-tuning and need to be extended to other styles.

\subsection{Visual Generative Algorithms}
 Hertzmann et al.~\shortcite{hertzmann2001image} introduce image analogies, a generative method based on a non-parametric texture model. Many recent approaches are based on CNNs  and use large datasets of input-output training image pairs to learn the mapping function \cite{gatys2015neural}. 
Inspired by the idea of variational autoencoders~\cite{kingma2013auto}, 
Johnson et al.~\shortcite{johnson2016perceptual} introduce the concept of perceptual loss to model the style transfer between paired dataset. 
Zhu et al. ~\shortcite{zhu2017unpaired} use generative adversarial networks to learn the mappings without paired training examples. These techniques have been used to  
generate natural images~\cite{karras2017progressive,sangkloy2017scribbler}, artistic images~\cite{li2017universal}, and videos~\cite{vondrick2016generating,li2018flow}.
Compared to previous visual generative methods, our approach can generate results of high resolution, can be applied to different painting media and is easy to extend to different painting media and artistic styles.

\subsection{Image Synthesis Using Machine Learning}
Many techniques have been proposed for image synthesis using machine learning.  \bl{Hu et al.~\shortcite{hu2018exposure} present a framework 
using reinforcement learning and generative adversarial network to learn photo post-processing.
}
Xie et al.~\shortcite{xie2012,xie2015stroke,xie2013personal} present a series of algorithms that simulate strokes using reinforcement learning and inverse reinforcement learning. These approaches learn a policy from either reward functions or expert demonstrations. 
In contrast to our algorithm, Xie et al.~\shortcite{xie2012,xie2015stroke,xie2013personal} focus on designing reward functions to generate orientational painting strokes. Moreover, their approach requires expert demonstrations for supervision. Ha et al.~\shortcite{DBLP:journals/corr/HaE17} collect a large-scale dataset of simple sketches of common objects with corresponding recordings of painting actions. Based on this dataset, a recurrent neural network model is trained in a supervised manner to encode and re-synthesize the action sequences. Moreover, the trained model is shown to be capable of generating new sketches. Following \cite{DBLP:journals/corr/HaE17}, Zhou et al.~\shortcite{zhou2018learning} use reinforcement learning and imitation learning to reduce the amount of supervision needed to train such a sketch generation model. 
In contrast to prior methods, \cite{jia2019paintbot} operate in a continuous action space with higher dimensions applying PPO\cite{schulman2017proximal} reinforcement learning algorithm to train the agent from scratch. It can handle dense images with high resolutions. 
\bl{We use the same painting environment as \cite{jia2019paintbot} to demonstrate the benefits of our proposed learning algorithm. Although both algorithms do not need imitation data from human experts,  self-supervised learning helps the reinforcement learning to converge to a better policy.}
Compared with prior visual generative methods, our painting agent can automatically generate results using a limited training dataset without paired dataset.

\subsection{Reinforcement Learning}
Reinforcement learning (RL) has achieved promising results recently in many problems, such as playing Atari games~\cite{mnih2013playing}, the game of Go~\cite{silver2017mastering} and robot control~\cite{levine2016end}. A major focus of this effort has been to achieve improved time and data efficiency of the learning algorithms. Deep Q-Learning has been shown to be effective for tasks with discrete action spaces~\cite{mnih2013playing}, and proximal policy optimization (PPO)~\cite{schulman2017proximal} is currently regarded as one of the most effective for continuous action space tasks. 
\bl{Hindsight experience replay \cite{andrychowicz2017hindsight} enables off-policy reinforcement learning to sample efficiently from rewards which are sparse and binary. \cite{andrychowicz2017hindsight} can be seen as a sampling approach for off-policy algorithms, while we treat self-supervised learning and reinforcement learning as two components. Compared with \cite{andrychowicz2017hindsight}, we present a practical approach to handle continuous space in a sparse reward setting and enhance the sampling efficiency by the self-supervised learning.}

\nocite{zheng2018strokenet,tang2018animated,xie2015stroke,xie2012,DBLP:journals/corr/HaE17}

\section{Self-Supervised Painting Agent}
In this section, we introduce notations, formulate the problem and present our self-supervised learning algorithm for natural media painting.

\begin{figure}
\centering
\includegraphics[width=0.43\textwidth]{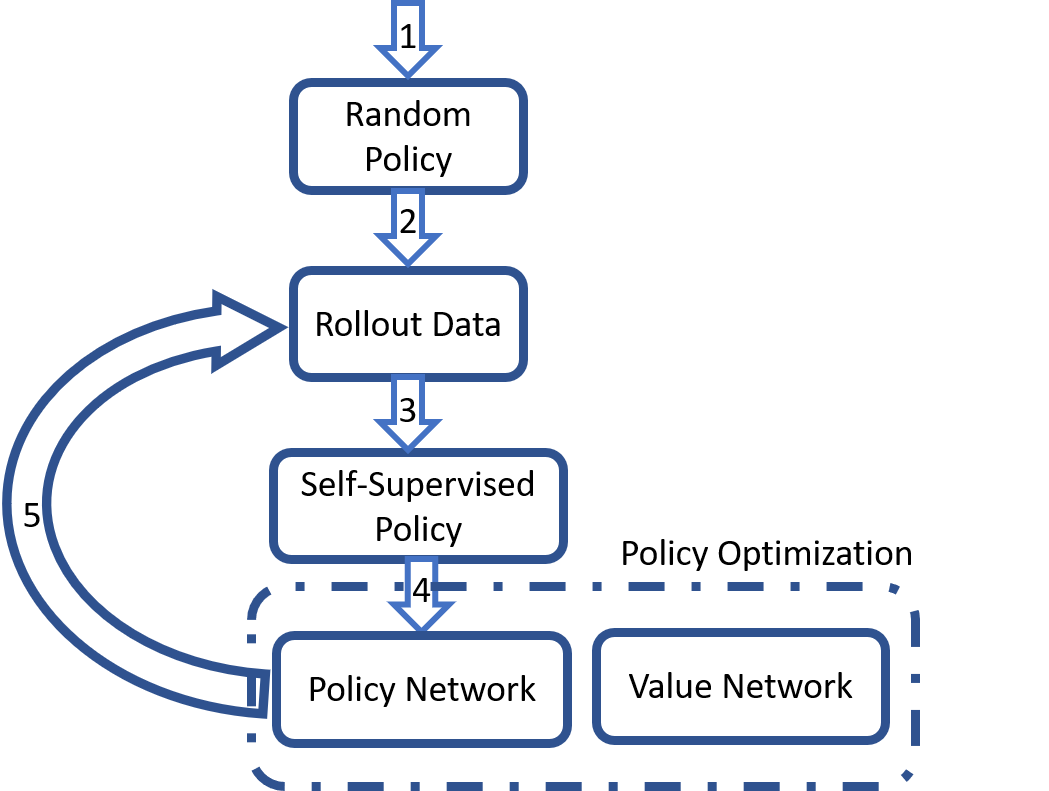}
\caption{\emph{Our Learning Algorithm:} 
We use self-supervised learning to generate paired dataset using a training dataset with reference images only and initialize the model for reinforcement learning.
Then we feed the trained policy network to self-supervised learning to generate the paired datasets with positive rewards.
(1) We initialize the policy network with random painting actions; 
(2) We rollout the policy by iteratively applying to the policy network to the painting environment to get paired data, followed by assigning the goal state $s^*$ as $\hat{s^*}$ and changing the rewards of each step accordingly; 
(3) We retrain the policy with the supervision data to generate the self-supervised policy, and use the behavior cloning to initialize the policy network; 
(4) We apply policy optimization \cite{schulman2017proximal} and update the policy; 
(5) We rollout the updated policy and continue the iterative algorithm. 
}
\label{fig:framework}
\vspace{-25px}
\end{figure}

\begin{table}
\begin{tabular}{ll}
\hline  
Symbol & Meaning   \\
\hline
$t$ & step index \\
$t_q$ & time steps to compute accumulated rewards \\
$s_t$ & current painting state of step $t$, canvas\\
$s^*$ & target painting state, reference image\\
$\hat{s^*}$ & reproduction of $s^*$ \\
$o_t$ & observation of step $t$ \\
$a_t$ & action of step $t$\\
$r_t$ & reward of step $t$ \\
$q_t$ & accumulated reward of step $t$ \\
$\gamma$ & discount factor for computing the reward \\
$\pi$ & painting policy, predict $a$ by $o$\\
$V_\pi$ & value function of the painting policy, \\
& predict $r$ by $o$ \\
$f(s)$ & feature extraction of state $s$ \\
$Render(a_t, s_t)$ & render function, render action to $s_t$\\
$Obs(s^*, s_t)$ & observation function, encode the current \\
 & state and the target state  \\
$Loss(s, s^*)$ & loss function, measuring distance between \\
    & state $s$ and objective state $s^*$\\
\hline
\end{tabular}
\label{Fig:param}
\caption{Notation and Symbols used in our Algorithm}
\vspace{-25px}
\end{table}

\subsection{Background}
\label{sec:background}
Self-supervised learning methods \cite{kolesnikov2019revisiting} are designed to enable learning without explicit supervision. The supervised signal for a pretext task is created automatically. 
It is a form of unsupervised learning where the data itself provides supervision. 
In its original formulation, this process is performed by withholding part of
the information of the data and training the classification or regression function to predict it.
The required task usually has a definition of the proxy loss so that it can be solved by self-supervised learning.
There are a variety of applications for self-supervised learning in different areas such as audio-visual analysis \cite{owens2018audio}, visual representation learning \cite{doersch2015unsupervised}, image analysis \cite{gidaris2018unsupervised}, robotics \cite{jang2018grasp2vec} etc.
\bl{In this paper, we use the term self-supervised learning to refer to the process of generating self-supervision data and using the data to initialize the policy network of the reinforcement learning framework.
}

\subsection{Problem Formulation}
Reproducing images with brush strokes can be formalized as finding a series of actions that minimizes the distance between the reference image and the current canvas in the desired feature space. Based on notations in Table 1, this can be expressed as minimizing the loss function:
\begin{equation}
    \label{eq:minimize}
    \pi^* = \argmin{Loss(\hat{s^*}, s^*)} 
\end{equation}
After we apply reinforcement learning to solve the problem by defining $Reward()$ function, we can get:
\bl{
\begin{equation}
    \pi^* = \argmax\sum_{t}^{N}{Reward(a_t, s_t)}
\end{equation}
}
\subsection{\bl{Painting Agent}}
In this section, we present the technical details of our reinforcement learning-based painting agent.

\subsubsection{Observation}
Our observation function is defined as follows. First, we encode the objective state (reference image) with the painting canvas. Second, we extract both the global and the egocentric view of the state. As mentioned in \cite{zhou2018learning, jia2019paintbot}, the egocentric view can encode the current position of the agent and it provides details about the state. The global view can provide overall information about the state. $o(s_i)$ is defined as Eq.\eqref{eq:obs}, given the patch size $(h_o, w_o)$ and the position of the brush position$(h_p, w_p)$.

\begin{equation}
\label{eq:obs}
\begin{split}
    o(s_i) = &\left\{ s_i\left[h_p-\frac{h_o}{2}:h_p+\frac{h_o}{2}, w_p-\frac{w_o}{2}:w_p+\frac{w_o}{2}\right]\right.,  s_i\\      &\left.s^*\left[h_p-\frac{h_o}{2}:h_p+\frac{h_o}{2}, w_p-\frac{w_o}{2}:w_p+\frac{w_o}{2}\right], s^* \right\}.
\end{split}
\end{equation}

\subsubsection{Action}
The action is defined as a vector in continuous space with positional information and paintbrush configurations. $a_ = \{dh, dw, width, color_R, color_G, color_B\}  \in \mathbb{R}^6$.  Each value is normalized to $[0, 1]$. The action is in a continuous space, which makes it possible to train the agent using policy gradient based reinforcement learning algorithms. 
The updated position of the paint brush after applying an action is computed by adding $(dh, dw)$ to the coordinates of the paint brush $(p_h', p_w') =(p_h+dh, p_w+dw)$.

\subsubsection{Loss Function}
The loss function defines the distance between the current state and the objective state. It can guide how the agent reproduces the reference image. In practice, we test our algorithm with $L_2$ defined as Eq.\eqref{eq:l2}, where $s$ is the image of size $h \times w \times c$.

\begin{eqnarray}
    &&L_2(s, s^*) = \frac{\sum^h_{i=1}\sum^w_{j=1}\sum^c_{k=1}||s_{ijk}-s^*_{ijk}||_2^2}{hwc}  \label{eq:l2}
\end{eqnarray}

For the self-supervised learning process, the loss function only affects reward computation. However, the reinforcement learning training process uses  $\{\hat{s}^*\}$ as the reference images to train the model and the loss function can affect the policy network.

\subsubsection{Policy Network}
To define the structure of the policy network, we consider the input as a concatenated patch of the reference image and canvas $82 \times 82 \times 3$ in egocentric view and global view, given the sample size of $41 \times 41 \times 3$. 
The first hidden layer convolves 64 $8 \times 8$ filters with stride 4, the second convolves 64 $4 \times 4$ filters with stride 2 and the third layer convolves 64 $3 \times 3$ filters with stride 1. After that, it connects to a fully-connected layer with 512 neurons. All layers use ReLU activation function \cite{krizhevsky2012imagenet}.\bl{For the training process, we add the criteria $r>0$ to expedite the training process. }

\subsubsection{Runtime Algorithm}
After we trained a model using self-supervised learning and reinforcement learning, we can apply the model to generate reference images with different resolutions.
First, we randomly sample a position from the canvas and draw a patch with size $(h_o, w_o)$ and feed it to the policy network. 
Second, we iteratively predict actions $a_t = \pi(o_t)$ and render them by environment until the value network $V_\pi$ returns a negative reward. 
Then we reset the environment by sampling another position from the canvas and keep the loop until $Loss(\hat{s^*}, s^*)$ less than $Thresh_{sim}$.

\subsection{Behavior Cloning}
Behavior cloning uses a paired dataset with observations and corresponding actions to train the policy to imitate an expert trajectory or behaviors. In our setup, the expert trajectory is encoded in the paired dataset $\{o^{(t)}, a^{(t)}\}$ which is related to step 4 in \autoref{fig:framework}.
We use behavior cloning to initialize the policy network of reinforcement learning with the supervised policy trained by paired data.
The paired dataset can be generated by a human expert or an optimal algorithm with global knowledge, which our painting agent does not have. 
Once the paired dataset $\{o^{(t)}, a^{(t)}\}$ is obtained, one solution is to apply supervised learning based on regression or classification to train the policy. The trained process can be represented using an optimization formulation as:
\begin{equation}
    \label{eq:problem}
    \pi^* = \argmin\sum_{t}^{N}{||\pi(o_t) - a_t ||}.
\end{equation}
It is difficult to generate such an expert dataset for our painting application because of the large variation in the reference images and painting actions. However, we can generate a paired dataset by rolling out a policy defined as Eq.\eqref{eq:rollout}, which can be seen as iteratively applying predicted actions to the painting environment.
For the painting problem, we can use the trained policy itself as the expert by introducing  self-supervised learning.

\subsection{Self-Supervised Learning}
\label{sec:self}
As we apply reinforcement learning to the painting problem, several new identities emerge as distinct from those of the classic controlling problems \cite{schulman2017proximal, schulman2015trust, mnih2013playing, sutton2000policy}. 
We use the reference image as the objective and encode it in the observation of the environment defined in Eq.\eqref{eq:obs}. As a result, the objective of the task Eq.\eqref{eq:problem} is not explicitly defined. Hence the rollout actions on different reference images $\{s^*\}$ can vary.

Through the reinforcement learning training process, the positive rewards in the high dimensional action space can be very sparse. In other words, only a small portion of actions sampled by policy network have positive rewards.
To change the reward distribution in the action space by increasing the probability of a positive reward, we propose using self-supervised learning. Our formulation uses the rollout of the policy as the paired data to train the policy network and retrains the model using reinforcement learning. 
Specifically, we replace the reference image $s^*$ with the final rendering of the rollout of the policy function $\hat{s^*}$. Moreover, we use the updated observation $\{\hat{o_t}\}$ and the actions $\{a_t\}$ as the paired supervised training dataset. 
\bl{For the rollout process of the trained policy $\pi$, we have:
\begin{eqnarray}
    &&a_t = \pi(o_{t-1}) \label{eq:rollout},   \\
    &&s_t = Render(s_{t-1}, a_t) \label{eq:render}, \\
    &&r_t =  \frac{Loss(s_{t-1}, s^*) - Loss(s_{t}, s^*)}{Loss(s_0, s^*)},\\
    &&o_t = Obs(s_t, s^*).
\end{eqnarray}
}
We can collect $\{o_{(t)}, a_{(t)}\}$ as the paired data. We denote the rendering of the final state as $\hat{s^*}$. 
The reward function is defined as the percentage improvement of the loss over the previous state.

Next, we modify $o_t$ and $r_t$ to a self-supervised representation as $\hat{o}_t$ and $\hat{r}_t$  as:
\begin{eqnarray}
    &&\hat{o}_t = Obs(s_t, \hat{s^*}), \\
    &&\hat{r}_t =  \frac{Loss(s_{t-1}, \hat{s^*}) - Loss(s_{t}, \hat{s^*})}{Loss(s_0, \hat{s^*})}, \\
    &&\hat{q}_t = \sum_{k=t}^{t_s}{\gamma^{k-t}r_k}.    
\end{eqnarray}
\bl{We use $\{\hat{o}_{(t)}, a_{(t)} , \hat{q}_{(t)}\}$ to train a self-supervised policy $\hat{\pi}$ and the value function $\hat{V_\pi}$. Algorithm \ref{alg:self} highlights the learning process for self-supervised learning. }

\begin{algorithm}[t]
  \caption{Self-Supervised Learning}
  \label{alg:self}
  \begin{algorithmic}[1]
    \REQUIRE 
     Set of objective states $\{s^{*(i)}\}$, its size is $n_s$
    \ENSURE Painting Policy $\pi$ and its value function $V_\pi$
    \FOR {$i = 1, \cdots, n_s$}
        \STATE $t=0$
        \STATE $s_0 = \textsc{Initialize}()$
        \STATE // Rollout the policy and collect the paired data with positive reward 
        \WHILE {$r \geq 0$} 
            \STATE $t = t+1$
	        \STATE $a_t = \pi(Obs(s_{t-1}, s^{*(i)}))$
	        \STATE $s_t = Render(s_{t-1}, a_t)$
	        \STATE $r = (Loss(s_{t-1},  s^{*(i)}) - Loss(s_{t},  s^{*(i)})))/Loss(s_0,  s^{*(i)}) $
	    \ENDWHILE
	    \STATE // Build self-supervised learning dataset
	    \bl{
	    \FOR {$j =0, \cdots, t-1$}
	        \STATE $\hat{o}_j = Obs(s_j, s_t)$
	        \STATE $\hat{r}_j = (Loss(s_{j+1},  s_t) - Loss(s_j,  s_t))/Loss(s_0,  s_t)$
	    \ENDFOR
	    
	    \STATE // Compute cumulative rewards
	    \FOR {$j =0, \cdots, t-1$}
	        \STATE $\hat{q}_j = \sum_{k=j}^{t-1}{\gamma^{k-j}\hat{r}_k}$
	    \ENDFOR 
	 \STATE $\pi=\textsc{Update}(\pi, \{\hat{o}_{(j)}, a_{(j)}\})$ // Initialize policy network for policy optimization
	 \STATE $V_\pi=\textsc{Update}(V_\pi, \{\hat{o}_{(j)}, \hat{q}_{(j)}, a_{(j)}\})$ // Initialize value network for policy optimization
	 }
	 \ENDFOR
  \bl{\RETURN $\pi$, $V_\pi$}
  \end{algorithmic}
\end{algorithm}



\section{Implementation}
\begin{figure}[h!]
\centering
\includegraphics[width=0.4\textwidth]{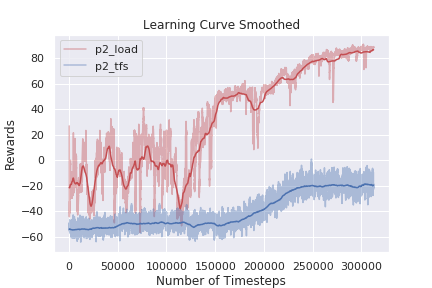}
\caption{\emph{Learning Curve Comparison} We evaluate our algorithm by plotting the learning curve of \bl{training from scratch} (blue) and \bl{training with self-supervised learning} (red). As shown in the figure, the method with self-supervision have better convergence and performance. }
\label{fig:vec}
\vspace{-15px}
\end{figure}

Our painting environment is similar to that in \cite{jia2019paintbot}, which is a simplified simulated painting environment. 
Our system can execute painting actions with parameters describing stroke size, color and positional information and updates the canvas accordingly.
We use a vectorized environment \cite{stable-baselines} for a parallel training process, as shown in \autoref{fig:vec}, to train our model.

\subsection{Performance}
In practice, we use a 16 core CPU and a GTX 1080 GPU to train the model with a vectorized environment of dimension 16. 
 We use SSPE \cite{jia2019paintbot} as $Render(a, s)$ to accelerate the training process. The learned policy can also be transferred to other simulated painting media like MyPaint or WetBrush \cite{chen2015wetbrush} to get different visual effects and styles.

\section{Results}
In this section, we highlight the results and compare the performance with prior learning-based painting algorithms.

For the first experiment, we apply a critic condition to reward each step $r_t \geq 0$ for $t \geq 5$. Once the agent fails the condition, the environment will stop the rollout.  We compare the cumulative reward $\sum_{t}{r_t}$ by feeding the same set of unseen images $\{s^{*(i)}\}$ to the environment. We use two benchmarks to test the generalization of the models. Benchmark1 is to reproduce an image $s^{*(i)}$ from a random image $s_j$. Benchmark2 is to reproduce an image $s^{*(i)}$ from a blank canvas. 
\bl{
It can lead to a higher cumulative reward of Benchmark2 because the initial loss of Benchmark1 is less than Benchmark2.
}
Each benchmark have 1000 $41\times41\times3$ patches.
As shown in Table 2, our combined training scheme outperforms using only self-supervised learning or only reinforcement learning. 
\begin{table}
\centering
\setlength{\tabcolsep}{3pt}
\begin{tabular}{|l|c|c|}
\hline
Benchmarks & Benchmark1 & Benchmark2 \\
\hline
Reinforcement Learning Only &  $4.67$ & $26.33$  	\\
\hline
Self-supervised Learning Only &  $31.20$ & $30.79$  	\\
\hline
Our Combined Scheme &  $49.42$ & $61.13$  \\
\hline
\end{tabular}
\caption{\emph{Comparison of Different Training Schemes:} We evaluate our method by comparing the average cumulative rewards on the test dataset.\bl{Self-supervised learning only refers to a policy that is trained with rollouts of a random policy by supervised learning, which reference image $s^*$ is replaced as the final rendering $s_t$. }
} 
\vspace{-25px}
\end{table}

For the second experiment, we evaluate the performance on the high-resolution reference images. We compute the $L_2$ loss and cumulative rewards and compare our approach with \cite{jia2019paintbot}. We draw $1000$ $400 \times 400$ patches from 10 reference images to construct the benchmark. Moreover, we iteratively apply both  the algorithms $1000$ times to reproduce the reference images. We use the same training dataset with images to train the models.
As shown in Table 3, our approach have a lower $L_2$, loss although both methods perform well in terms of cumulative rewards.

\begin{table}
\centering
\setlength{\tabcolsep}{3pt}
\begin{tabular}{|l|c|c|}
\hline
Approaches & Cumulative Rewards & $L_2$ Loss \\
\hline
PaintBot \cite{jia2019paintbot} &  $97.74$ & $1920$  	\\
\hline
LPaintB  &  $98.25$ & $1485$  \\
\hline 
\end{tabular}
\label{tab:compare2}
\caption{\emph{Comparison with Previous Work} We evaluate our method by comparing the average cumulative reward  and $L_2$ loss between final rendering and the reference image Eq.\eqref{eq:l2} on the test dataset.
} 
\vspace{-15px}
\end{table}

\begin{figure}[h]
\centering
\includegraphics[width=0.45\textwidth]{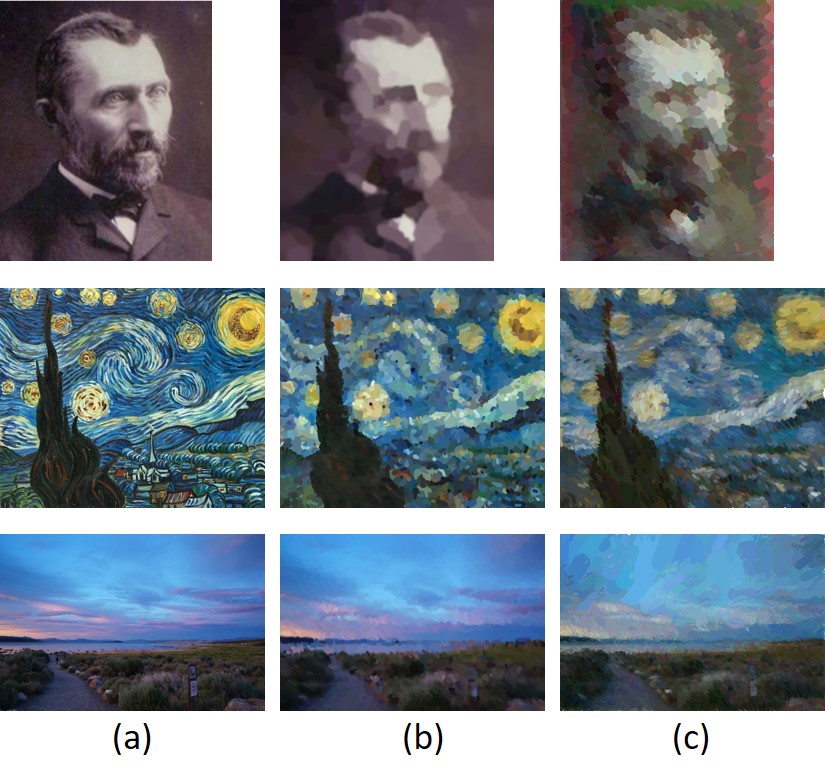}
\caption{\emph{Our results compared with \cite{jia2019paintbot}} 
 We compare the final rendering result using the same scale of the reference image and the same amount of painting actions.
 (a) are the reference images. (b) are generated by our painting agent (c) are generated by the agent \cite{jia2019paintbot}. \bl{We demonstrate the benefits of self-supervised learning by reference images with different resolutions. The training dataset for both algorithms consists of 374 $41 \times 41 \times 3$ patches sampling from one painting.
 }
 }
\label{fig:test}
\end{figure}

{\text{\ }}


\section{Conclusion}
We present a novel approach for stroke-based image reproduction using self-supervised learning and reinforcement learning.
Our approach is based on a feedback loop with reinforcement learning and self-supervised learning. 
We  modify and reuse the rollout data of the previously trained policy network and feed it into the reinforcement learning framework. 
We compare our method with both the model trained with only self-supervised learning and the model trained from scratch by reinforcement learning.
The result shows that our combination of self-supervised and reinforcement learning can greatly improve efficiency of sampling and performance of the policy. 

One major limitation of our approach is that the generalization of the trained policy is highly dependent on the training data. 
Although reinforcement learning enables the policy to generalize to different states that supervised learning cannot address, the states still depend on the training data. 
Specifically, the distribution of generated supervision data is not close to the unseen data.
\bl{Another limitation is the result generated by our method is not sharp enough, especially for the high contrast regions of the reference image. It can be improved by increasing either the total number of strokes or the resolutions of reference images but we still need a better definition of reward/loss to the problem.}

For future work, we aim to enlarge the runtime steps and action space of the painting environment so that the data generated by self-supervised learning can be closer to the distribution of the unseen data. 
Our current setup includes most common stroke parameters like brush size, color, and position, but the painting parameters describing pen tilting, pen rotation, and pressure are not used.
\bl{Moreover, we aim to develop a better definition of reward/loss to mitigate the blurry effects.}

\footnotesize{
\bibliographystyle{eg-alpha}
\bibliography{ref}
}
\end{document}